\documentclass{INTERSPEECH2023}

\usepackage{epsfig}
\usepackage{epstopdf, dblfloatfix, subfig}
\usepackage{balance}
\usepackage{multirow, multicol}
\usepackage{soul}
\usepackage{tikz}
\usetikzlibrary{calc}

\makeatletter
\newif\if@anonymize

\@anonymizefalse  

\if@anonymize
  \newcommand{\highlight@DoHighlight}{
    \fill [outer sep = -15pt, inner sep = 0pt, color=black]
          ($(begin highlight)+(0,8pt)$) rectangle ($(end highlight)+(0,-3pt)$) ;
  }

  \newcommand{\highlight@BeginHighlight}{
    \coordinate (begin highlight) at (0,0) ;
  }

  \newcommand{\highlight@EndHighlight}{
    \coordinate (end highlight) at (0,0) ;
  }

  \newdimen\highlight@previous
  \newdimen\highlight@current
  \newlength{\item@width}

  \DeclareRobustCommand*\anonymize{%
    \SOUL@setup
    \def\SOUL@preamble{%
      \begin{tikzpicture}[overlay, remember picture]
        \highlight@BeginHighlight
        \highlight@EndHighlight
      \end{tikzpicture}%
    }%
    \def\SOUL@postamble{%
      \begin{tikzpicture}[overlay, remember picture]
        \highlight@EndHighlight
        \highlight@DoHighlight
      \end{tikzpicture}%
    }%
    \def\SOUL@everyhyphen{%
      \discretionary{%
        \SOUL@setkern\SOUL@hyphkern
        \SOUL@sethyphenchar
        \tikz[overlay, remember picture] \highlight@EndHighlight ;%
      }{%
      }{%
        \SOUL@setkern\SOUL@charkern
      }%
    }%
    \def\SOUL@everyexhyphen##1{%
      \SOUL@setkern\SOUL@hyphkern
      \settowidth{\item@width}{##1}%
      \makebox[\item@width]{}%
      \discretionary{%
        \tikz[overlay, remember picture] \highlight@EndHighlight ;%
      }{%
      }{%
        \SOUL@setkern\SOUL@charkern
      }%
    }%
    \def\SOUL@everysyllable{%
      \begin{tikzpicture}[overlay, remember picture]
        \path let \p0 = (begin highlight), \p1 = (0,0) in \pgfextra
          \global\highlight@previous=\y0
          \global\highlight@current =\y1
        \endpgfextra (0,0) ;
        \ifdim\highlight@current < \highlight@previous
          \highlight@DoHighlight
          \highlight@BeginHighlight
        \fi
      \end{tikzpicture}%
      \settowidth{\item@width}{\the\SOUL@syllable}%
      \makebox[\item@width]{}%
      \tikz[overlay, remember picture] \highlight@EndHighlight ;%
    }%
    \SOUL@
  }
\else
  \newcommand{\anonymize}[1]{#1}
\fi
\makeatother

\usepackage{color}
\newcount\Comments  
\definecolor{darkgreen}{rgb}{0,0.5,0}
\definecolor{darkred}{rgb}{0.7,0,0}
\definecolor{teal}{rgb}{0.1,0.6,0.7}
\definecolor{blue}{rgb}{0.0,0.1,0.9}
\definecolor{orange}{rgb}{1.,0.7,0.0}
\definecolor{lightblue}{rgb}{0.70, 0.80, 0.89}
\definecolor{pink}{rgb}{0.3,0,0}
\definecolor{yellow}{rgb}{0,0.5,0.5}

\newcommand{\kibitz}[2]{\ifnum\Comments=1{{\textcolor{#1}{\textsf{\footnotesize [#2]}}}}\fi}

\title{The 2022 NIST Language Recognition Evaluation}
\name{{Yooyoung Lee$^1$, Craig Greenberg$^1$, Eliot Godard$^{1,*}$, Asad A. Butt$^{1,*}$, Elliot Singer$^2$, Trang Nguyen$^2$, Lisa Mason$^3$, Douglas Reynolds$^3$} \thanks{$^*$NIST Associates}}
\address{
  $^1$NIST ITL/IAD/Multimodal Information Group, MD, USA\\
  $^2$MIT Lincoln Laboratory, Lexington, MA, USA\\
  $^3$U.S. Department of Defense, MD, USA}
\email{lre\_poc@nist.gov}

\graphicspath{{./figs/}}

\begin{document}

\setlength{\abovedisplayskip}{3pt}
\setlength{\belowdisplayskip}{3pt}
\setlength{\abovecaptionskip}{5pt}

\maketitle
\begin{abstract}
  In 2022, the U.S. National Institute of Standards and Technology (NIST) conducted the latest Language Recognition Evaluation (LRE) in an ongoing series administered by NIST since 1996 to foster research in language recognition and to measure state-of-the-art technology. Similar to previous LREs, LRE22 focused on conversational telephone speech (CTS) and broadcast narrowband speech (BNBS) data. LRE22 also introduced new evaluation features, such as an emphasis on African languages, including low resource languages, and a test set consisting of segments containing between 3s and 35s of speech randomly sampled and extracted from longer recordings. A total of 21 research organizations, forming 16 teams, participated in this 3-month long evaluation and made a total of 65 valid system submissions to be evaluated. This paper presents an overview of LRE22 and an analysis of system performance over different evaluation conditions. The evaluation results suggest that Oromo and Tigrinya are easier to detect while Xhosa and Zulu are more challenging. A greater confusability is seen for some language pairs. When speech duration increased, system performance significantly increased up to a certain duration, and then a diminishing return on system performance is observed afterward. 

  
\end{abstract}

\section{Introduction}\label{sec:intro}
\begin{figure}[t]
	\centering
	\includegraphics[width=0.8\linewidth, clip, trim=0mm 0mm 0mm 0mm]{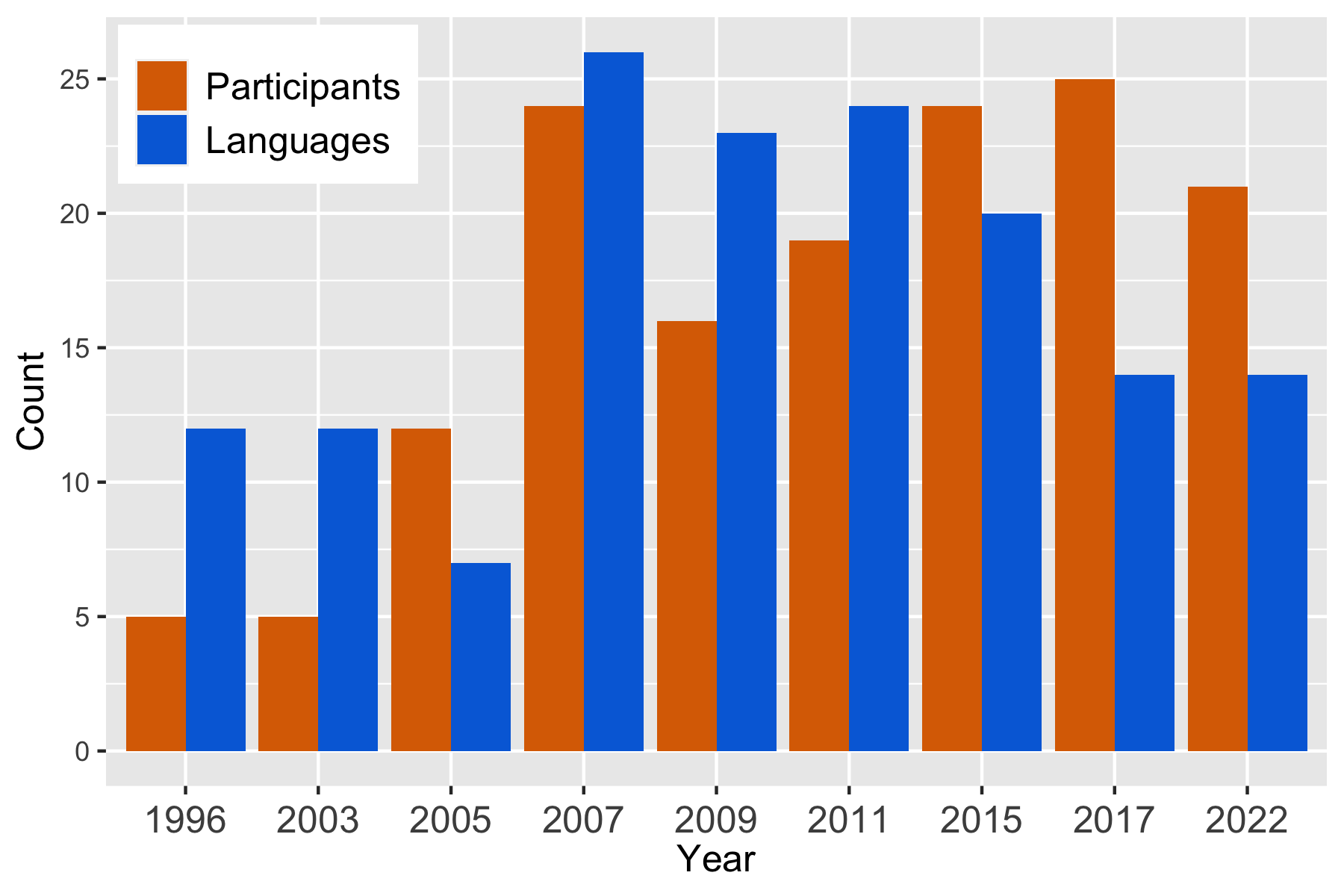}
	\caption{Language and participant count for the NIST LREs}
	\label{fig:par_stat}
	\vspace{-3mm}
\end{figure}

The 2022 NIST Language Recognition Evaluation (LRE), held in fall of 2022, was the latest in an ongoing series of language recognition evaluations conducted by NIST since 1996~\cite{nistlre}. The primary objectives of the LRE series are to: 1) advance language recognition technologies with innovative ideas, 2) facilitate the development of language recognition technology by providing data and research direction, and 3) measure the performance of the current state-of-the-art technology. Figure~\ref{fig:par_stat} shows the number of target languages and participants (based on sites) for all NIST LREs.

\begin{figure}[t]
	\centering
	\includegraphics[width=\linewidth, clip, trim=0mm 0mm 0mm 0mm]{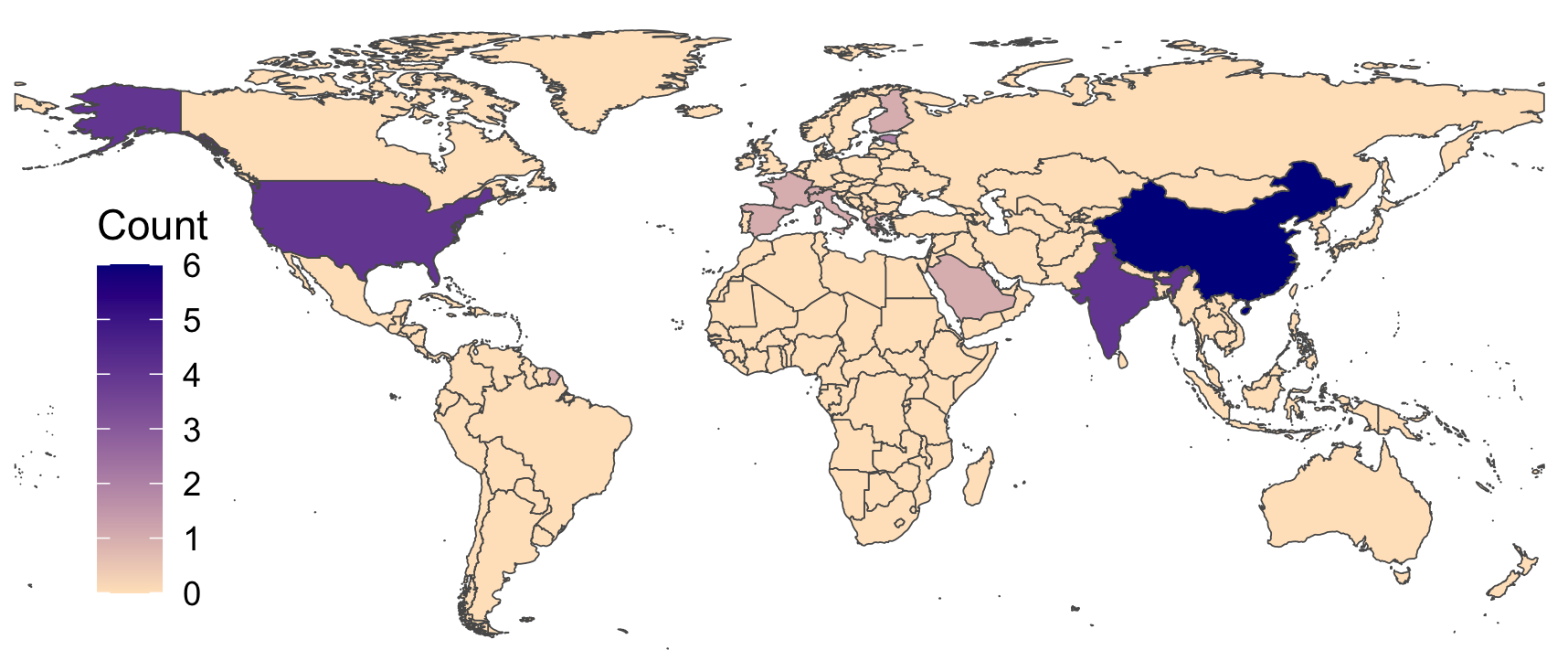}
	\caption{Heatmap of the world showing the number of LRE22 participating sites per country.}
	\label{fig:map}
	\vspace{-4mm}
\end{figure}

LRE22 was conducted entirely online using a web-based platform like LRE15~\cite{lre2015} and LRE17~\cite{lre2017,sadjadi18_interspeech}. The updated LRE22 web-platform\footnote{\anonymize{https://lre.nist.gov}} supported a variety of evaluation activities, such as registration, data license submission, data distribution, system output submission and validation/scoring, and system description/presentation uploads. A total of 16 teams from 21 organizations in 13 different countries made submissions for LRE22. Figure~\ref{fig:map} displays a world map with heatmap representing the number of participating sites per country. Since two teams did not submit valid system descriptions, analysis considering only 14 teams in presented this paper. It should be noted that all participant information, including country, was self-reported.




\section{Task}
\label{sec:task}
The general task in the NIST LREs is language detection, i.e. to automatically determine whether a particular target language was spoken in a given test segment of speech. Since LRE11~\cite{martin_odyssey14}, the focus of the language detection task had turned to distinguishing between closely related, and sometimes mutually intelligible, languages. However LRE22 introduced a new emphasis on distinguishing between African languages, including low resource languages. Table~\ref{tbl:tgt_langs} shows the 14 target languages included in LRE22. Similar to LRE17, LRE22 participants were required to provide a 14-dimensional vector of log-likelihood scores corresponding to the languages in Table~\ref{tbl:tgt_langs}. Unlike LRE17, language clusters were not considered in this evaluation; a language cluster is a group of two or more consonant sounds with those from the same speech community~\cite{Ahmad}.

Like LRE17, there were two training conditions in LRE22: \textit{fixed} and \textit{open}. For the \textit{fixed} training condition, participants were restricted to use only a limited pre-specified set of data for system training and target model development. For the \textit{open} training condition, participants were allowed to utilize unlimited amounts of publicly available and/or proprietary data for their system training and target model development. To facilitate more meaningful cross-system comparisons, LRE22 participants were required to provide submissions to the \textit{fixed} condition while participation in the optional \textit{open} condition was strongly encouraged to understand the impacts that larger amounts of training and development data have on system performance.
In order to encourage participation in the \textit{open} training condition, the deadline for this condition was made one week later than the required \textit{fixed} training condition submission deadline. A total of 65 valid submissions were received, 40 for the \textit{fixed} training condition and 25 for the \textit{open} condition. LRE participants were required to specify one submission as \textit{primary} for each training condition they took part in, while all other systems submitted were considered \textit{alternate}.

\begin{table}[]
		\begin{center}
            \caption{LRE22 target languages}
			\begin{tabular}{|p{2.8cm}|p{1cm}||p{1.15cm}|p{1.15cm}|}
				\hline
				\textbf{Language} & \textbf{Code} & \textbf{Language} & \textbf{Code} \\ 
				\hline
				Afrikaans & afr-afr & Ndebele & nbl-nbl \\
				\hline
                Tunisian Arabic &  ara-aeb & Oromo & orm-orm\\
				\hline
                Algerian Arabic &  ara-arq & Tigrinya & tir-tir\\
				\hline
                Libyan Arabic &  ara-ayl & Tsonga & tso-tso\\
				\hline
                South African English &  eng-ens & Venda & ven-ven\\
				\hline
                Indian-accent South African English & eng-iaf & Xhosa & xho-xho\\
				\hline
                North African French &  fra-ntf & Zulu & zul-zul\\
				\hline
			\end{tabular}
			\label{tbl:tgt_langs}
		\end{center}
\vspace{-8mm}
\end{table}



\section{Data}
\label{sec:data}

This section provides a brief description of data used in LRE22 for training, development (\textit{dev}), and evaluation (\textit{test}) sets, along with the associated metadata. 

\subsection{Training set}
As mentioned in Section \ref{sec:task}, there were two training conditions in LRE22. The \textit{fixed} condition limited the system training and development data to the following specific data sets provided to participants by the Linguistic Data Consortium (LDC): 2017 NIST LRE \textit{dev} set and previous NIST LRE training data (LDC2022E16), 2017 NIST LRE \textit{test} set (LDC2022E17), 2022 NIST LRE \textit{dev} set (LDC2022E14). The VoxLingua107 data set \cite{voxlingua} was also permitted for use in the \textit{fixed} condition. The \textit{open} training condition removed the limitations of the \textit{fixed} condition. In addition to the data listed in the \textit{fixed} condition, participants could use any additional data to train and develop their system, including proprietary data and data that are not publicly available. LDC also made selected data from the IARPA Babel Program~\cite{babeldata} available to participants to be used in the \textit{open} training condition.

\subsection{Development and test sets}
The development (\textit{dev}) set is normally used to build/optimize a system model during the development process while the evaluation (\textit{test}) set is used to evaluate the performance of the system model. The speech segments in the LRE22 \textit{dev} and \textit{test} sets were selected from data sets collected by the Linguistic Data Consortium (LDC) to support LR technology evaluations; namely the Maghrebi Language identification (MAGLIC), Speech Archive of South African Languages (SASAL), and Low Resource African Languages (LRAL) corpora.
The MAGLIC corpus was a CTS-only collection based in Tunisa and includes four regional language varieties spoken in North Africa: Algerian Arabic, Libyan Arabic, Tunisian Arabic, and North African French.
The SASAL corpus was a CTS and BNBS collection located in South Africa and contains several African language varieties, a subset of which were included in LRE22: Afrikaans, Ndebele, Tsonga, Venda, Xhosa, and Zulu, as well as South African English and Indian-accented South African English.
The LRAL corpus was a BNBS collection based in Ethiopia, and, of the languages in LRAL, two were selected for inclusion in LRE22: Oromo and Tigrinya. 

\begin{figure}[!ht]
	\centering
	\includegraphics[width=0.8\linewidth, clip, trim=4mm 10mm 1mm 12mm]{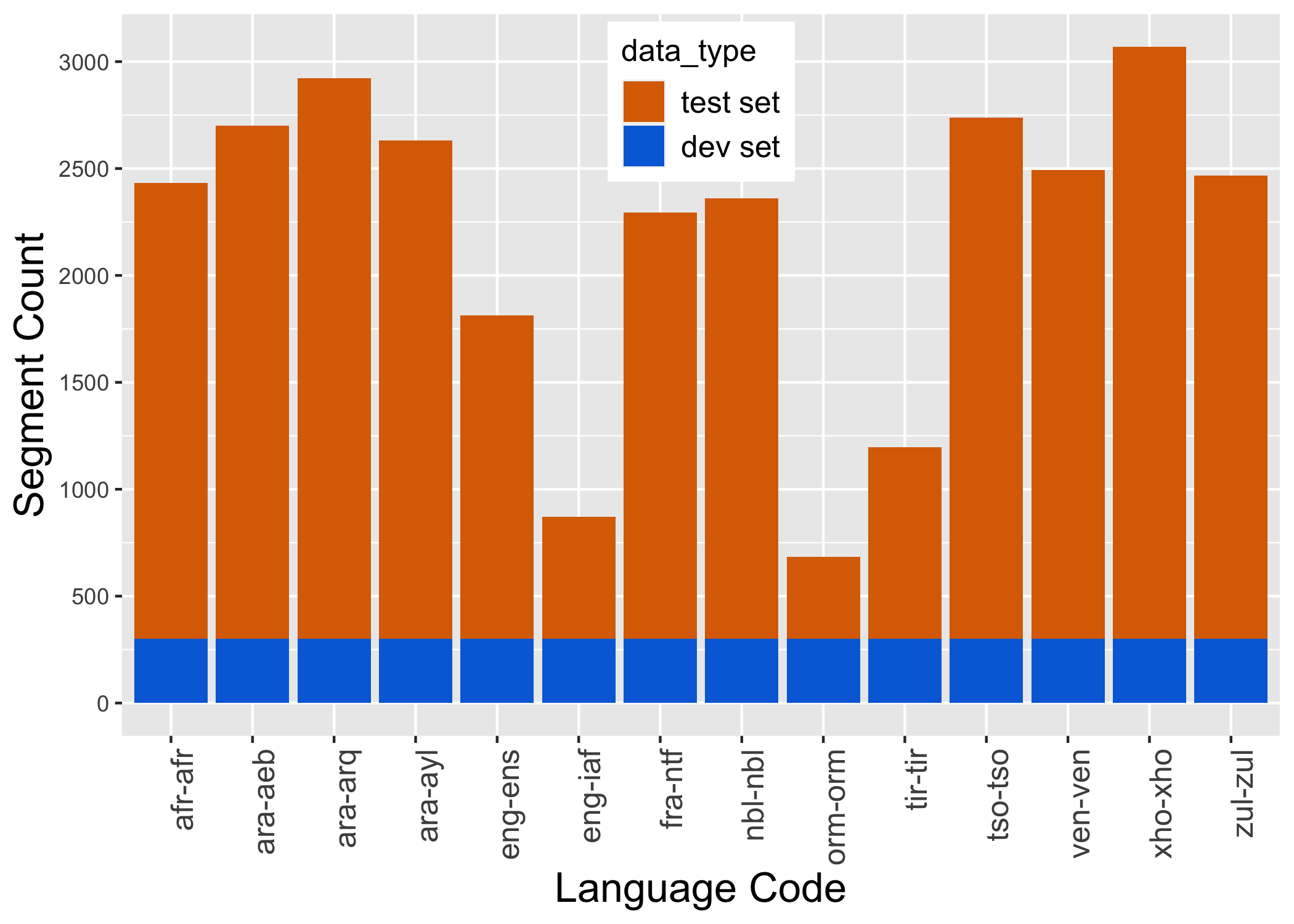}
	\caption{Distribution of speech segments per target language for both \textit{dev} and \textit{test} sets}
	\label{fig:dev_test_set_dis}
\end{figure}
\vspace{-7pt}

All audio data provided was sampled at 8~kHz, a-law encoded, and formatted as SPHERE \cite{sphere2012} files. When the source audio recordings were higher bandwidth or encoded differently, they were downsampled and transcoded to 8-kHz a-law. Unlike in previous LREs, the amount of speech in the LRE22 segments was uniformly sampled between approximately 3 and 35 seconds, as determined by an automatic speech activity detector. Figure~\ref{fig:dev_test_set_dis} shows a stacked histogram for the \textit{dev} and \textit{test} sets. The \textit{dev} set consisted of 300 segments per target language while the \textit{test} set contained a total of 26,473 segments ranging from 383 to 2,769 segments across the target languages.


\begin{figure}[!ht]
\centering
  \includegraphics[width=0.8\linewidth]{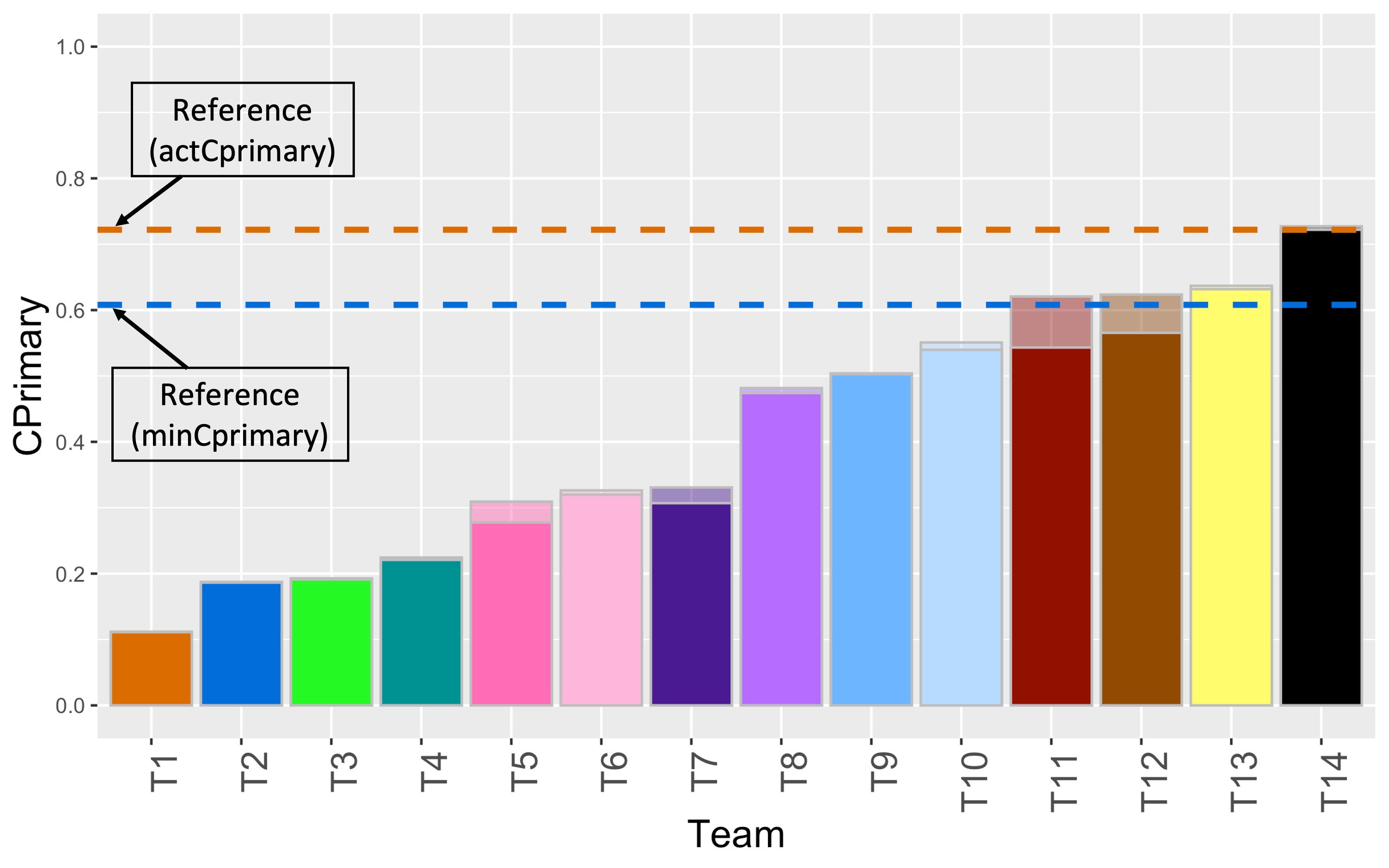}
  \caption{System performance (actual and minimum costs) on primary submissions under the \textit{fixed} training condition}
  \label{fig:fixed_p}
   \vspace{-3mm}
\end{figure}

\begin{figure}[!ht]
\centering
  \includegraphics[width=0.8\linewidth]{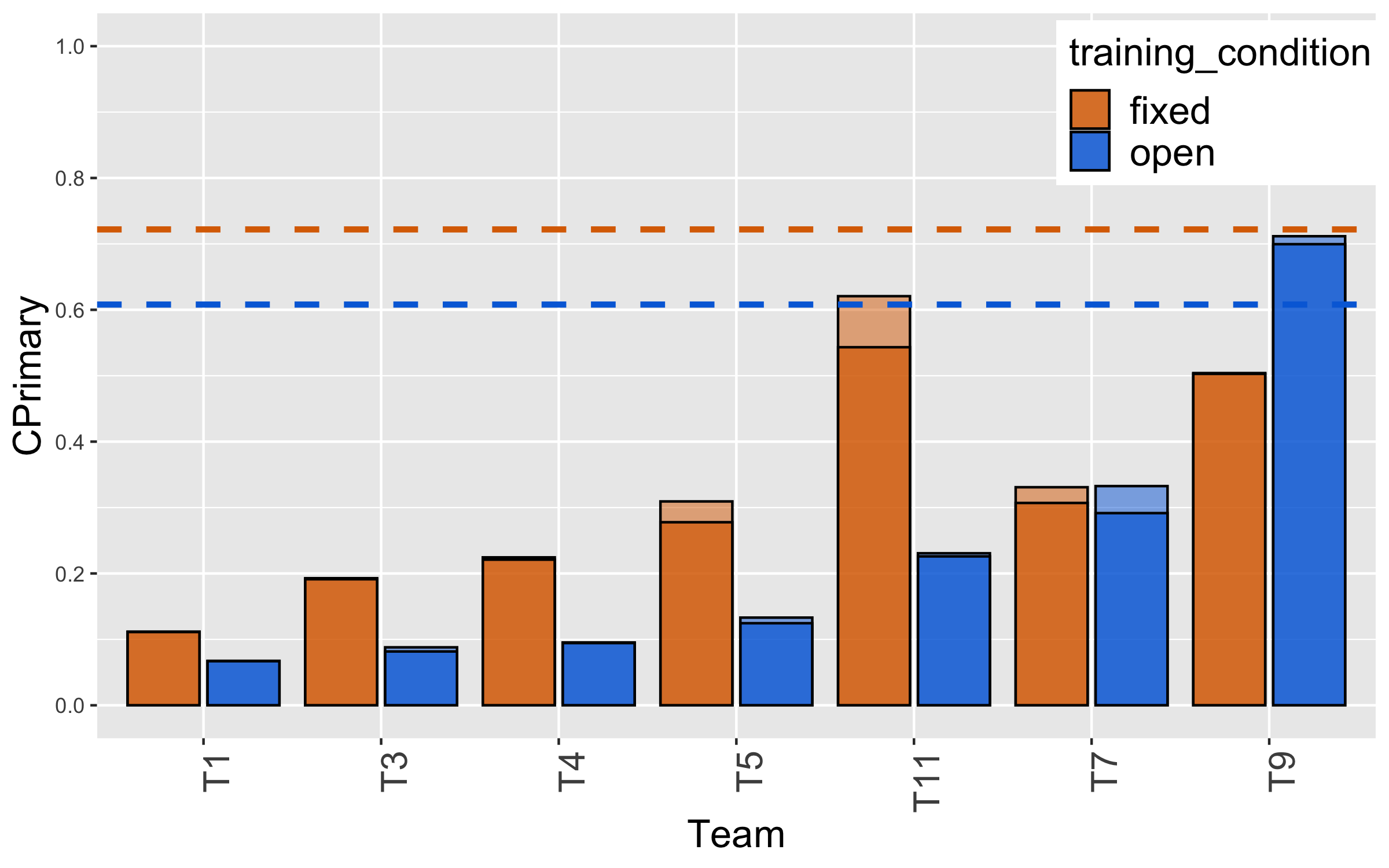}
  \caption{A performance comparison of the \textit{fixed} and \textit{open} training conditions}
  \label{fig:fixed_vs_open}
   \vspace{-4mm}
\end{figure}

\subsection{Metadata}

The metadata collected by LDC can be categorized into audio- and audit-related metadata. The audio metadata indicates information related to the audio recording or segment, such as speech duration, data source type (i.e., either CTS or BNBS), and source file (i.e., the original recording from which the audio segment was extracted). 
The audit metadata reflects a human auditor’s judgement of the speech, having listened to an audio recording, such as whether the recording contained a single speaker, if the person speaking was a native speaker, the speech clarity, the speaker sex, or if the recording took place in a noisy environment. In this paper, we limit our analyses on data source type and speech duration.

\section{Performance Measure}
\label{sec:task_metric}
As stated in the Section~\ref{sec:task}, LRE22 participants were required to provide a 14-dimensional vector of log-likelihood scores for the 14 target languages (see Table~\ref{tbl:tgt_langs} for the LRE22 target languages). Unlike LRE17, language clusters were not considered in this evaluation.
Pair-wise performance was computed for all target/non-target language pairs. A decision threshold derived from log-likelihood ratios was used to determine the number of missed detections and false alarms, computed separately for each target language. 
The missed detections (Misses) indicate the segments that are the target language, but are not predicted to be, while the false alarms (FAs) indicate the segments that are falsely identified as the target language. 
The probabilities of missed detections ($P_{Miss}$) and false alarms ($P_{FA}$) are then combined using a linear cost function~\cite{lre2022}:
\begin{align}
C(L_T,L_N) = 
&C_{Miss} \times P_{Target} \times P_{Miss}(L_T) + \nonumber \\
&C_{FA}  \times  (1-P_{Target}) \times P_{FA}(L_T,L_N)
\label{eqn:cdet}
\end{align}
where $L_T$ and $L_N$ are target and non-target languages, respectively. Here, $C_{Miss}$ (cost of a missed detection), $C_{FA}$ (cost of a false alarm), and $P_{Target}$ (the \textit{a priori} probability of the specified target language) are application-motivated cost model parameters. 
Two sets of cost-function parameters were used in LRE22: the first set of parameters provides equal weighting to the costs of errors ($C_{Miss}=C_{FA}=1$) and a target probability of 0.5, while the second set of parameters changed the target probability to 0.1.
The final metric, $C_{Primary}$, consisted of the mean value of the costs using the two different cost function parameters, normalized by dividing by the cost of a ``no information'' system.
Costs using thresholds that minimize the Bayes risk, $actC_{Primary}$, as well as using thresholds that minimize the empirical cost, $minC_{Primary}$, were computed. 
We refer readers to the LRE22 evaluation plan \cite{lre2022} for details of the performance measures.

\section{Results and Analyses}
A total of 14 teams from academic and industrial sectors successfully completed LRE22. For both the \textit{fixed} and \textit{open} training conditions, the teams were allowed to have one \textit{primary} submission and one or more \textit{alternate} submissions. In this section, we present a summary of results and key findings on the \textit{primary} submissions using the performance metrics defined in Section~\ref{sec:task_metric}.

Figure~\ref{fig:fixed_p} illustrates system performance for all the \textit{primary} submissions under the \textit{fixed} training condition. The x-axis are anonymized team names and the y-axis are $C_{Primary}$ values for both the actual and minimum costs (N.B., a lower $C_{Primary}$ value indicates better performance). The orange dashed-line indicates an actual cost, $actC_{Primary}$, and the blue is a minimum cost, $minC_{Primary}$, for a reference system; we used an off-the-shelf algorithm as a reference to validate the LRE22 data construction and evaluation process. The reference system was trained and fine-tuned only on VoxLingua107 and the LRE22 development set. The shaded color on each team’s bar indicates the difference between $actC_{Primary}$ and $minC_{Primary}$, which indicicates a calibration error. In Figure~\ref{fig:fixed_p}, we observe that, given the primary submissions under the \textit{fixed} condition, the $C_{Primary}$ values range from 0.11 to 0.73 across all the teams. It is observed that the top-performing systems (e.g., T1-T4) have small calibration errors (i.e., the absolute difference between the actual and minimum costs is relatively small) while a few teams (e.g., T5, T7, T11 and T12) are less well-calibrated.



\begin{figure}[!ht]
	\centering
	\includegraphics[width=0.95\linewidth, clip]{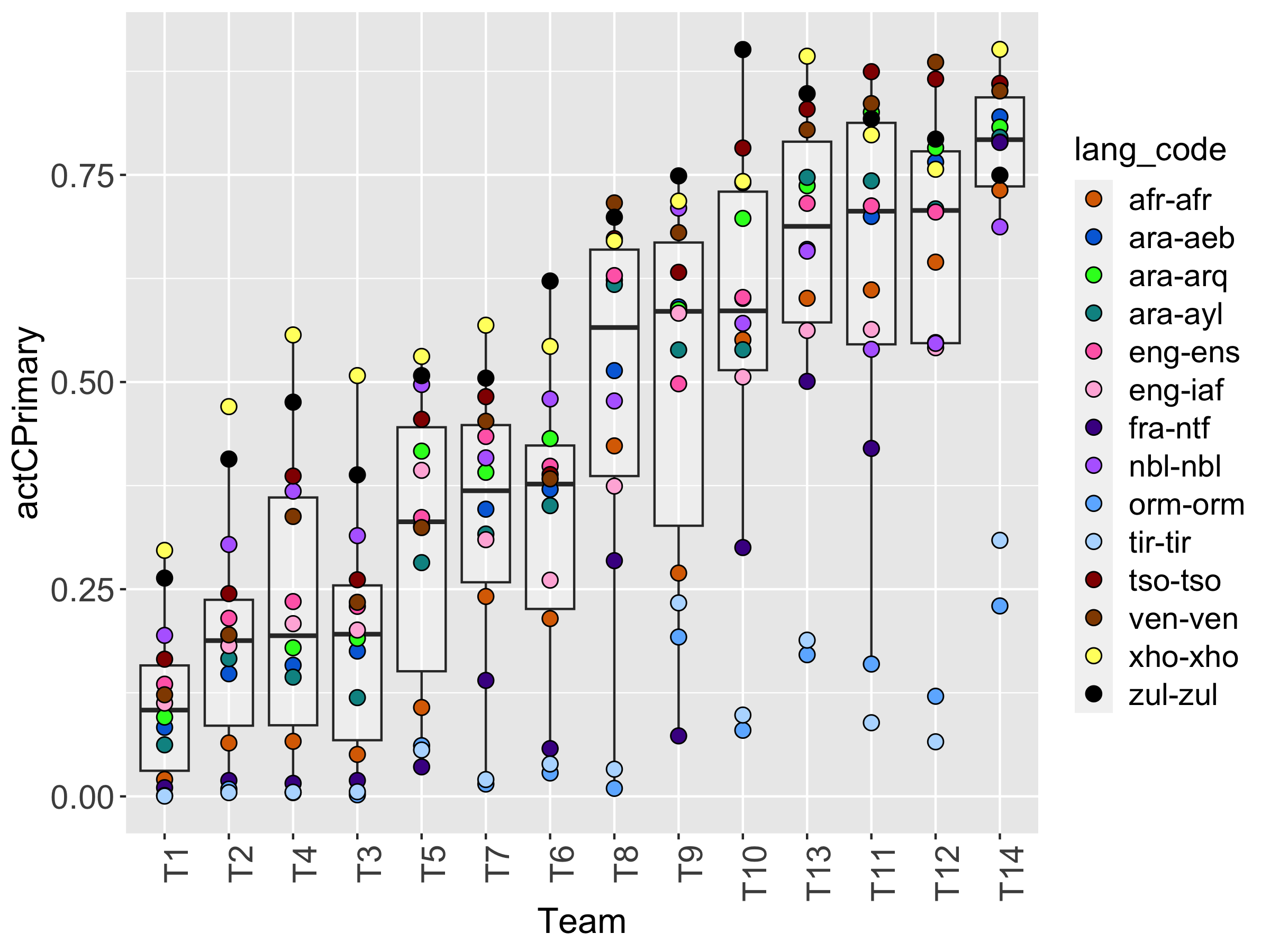}
	\caption{A language-level performance on primary submissions under the \textit{fixed} training condition}
	\label{fig:lang_level_fixed_p}
  \vspace{-4mm}
\end{figure}


As described in Section~\ref{sec:task}, the \textit{fixed} training condition is required while \textit{open} is optional; 7 out of the 14 teams submitted their system outputs to the \textit{open} training condition. Figure~\ref{fig:fixed_vs_open} illustrates a performance comparison of training conditions (\textit{fixed} vs \textit{open}) for the seven teams only (ordered by \textit{open} system performance). The result shows that system performance from the \textit{open} condition generally outperforms the \textit{fixed} condition submission across the teams (except T9), and a calibration error is observed in team T7 under the \textit{open} training condition.

To understand variability of language-level system performance and language detection difficulty, Figure~\ref{fig:lang_level_fixed_p} illustrates a box plot of the primary submission performance under the \textit{fixed} training condition. The x-axis is a team name (ordered by median), the y-axis is the actual cost ($actC_{Primary}$), and each point represents a target language. The black line within a box is the median, the box edges represent the lower quartile and upper quartile, and the whiskers extending from the box indicate variability outside the upper and lower quartiles. We observe a high dispersion of language performance for a few teams such as T4, T5, and T9. Overall, the \textit{Oromo (orm-orm)} and \textit{Tigrinya (tir-tir)} points marked in blue are located in the bottom side of Figure~\ref{fig:lang_level_fixed_p} (easier to detect) while \textit{Xhosa (xho-xho)} and \textit{Zulu (zul-zul)} are in the top (harder to detect); a similar trend is observed across the teams. 



\begin{figure}[!ht]
        \centering
	\includegraphics[width=\columnwidth, clip]{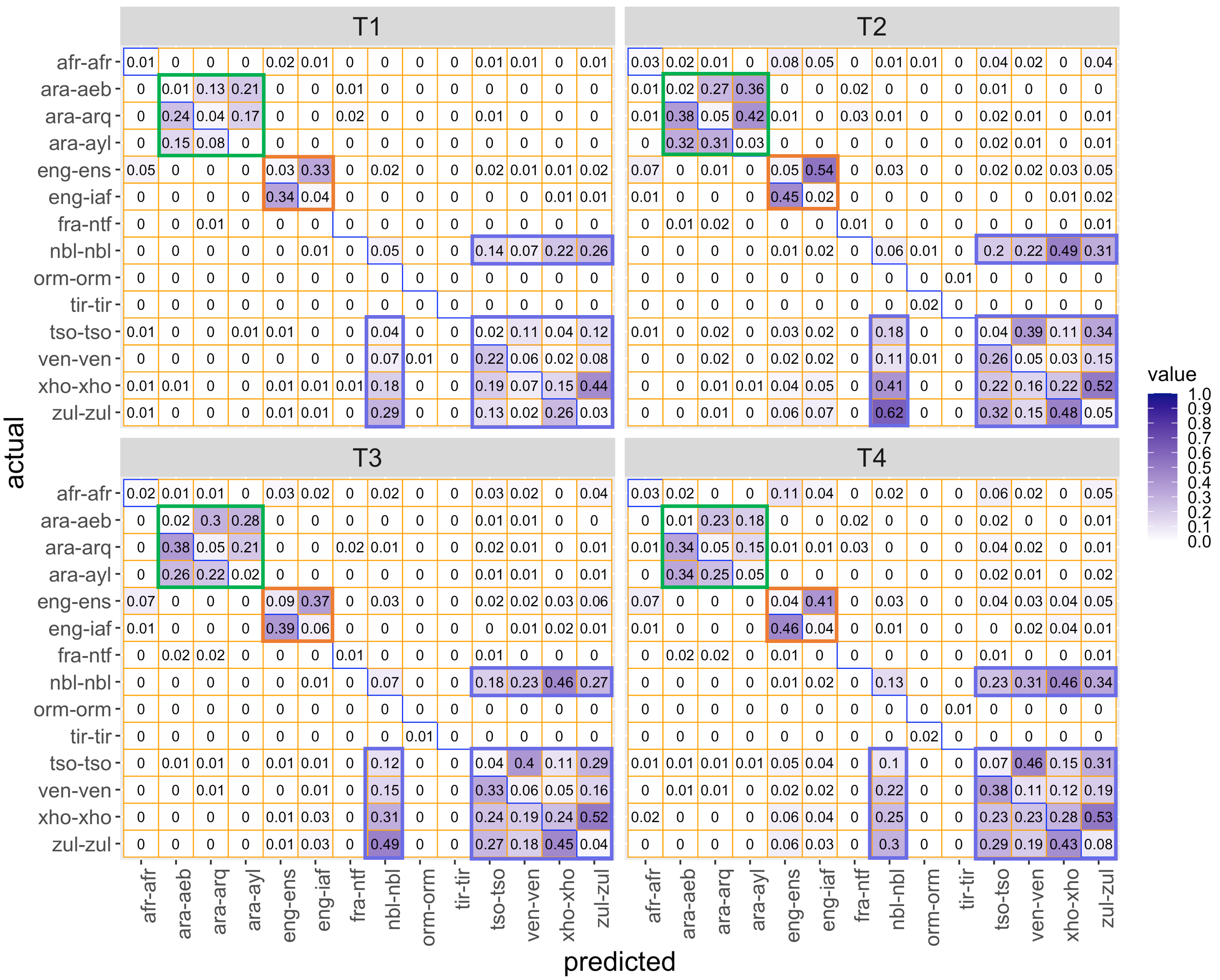}
	\caption{Language confusability of the leading systems}
	\label{fig:confusion_heatmap}
	\vspace{-2mm}
\end{figure}

\begin{figure}[!ht]
	\centering
	\includegraphics[width=0.9\linewidth, clip]{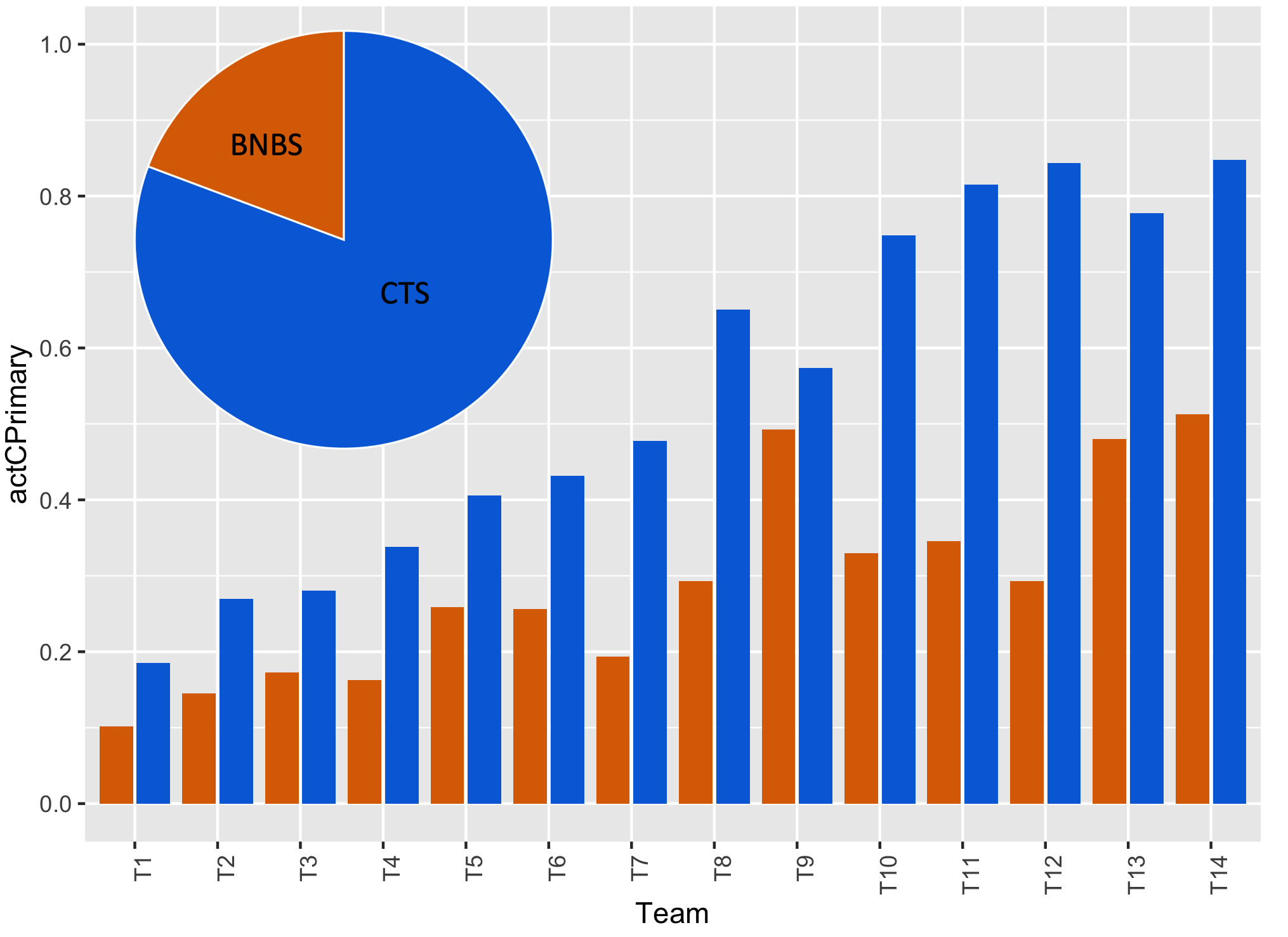}
	\caption{\textit{A data source type} distribution and effect on systems}
	\label{fig:partiton_data_source}
  \vspace{-3mm}
\end{figure}


\begin{figure}[!ht]
\vspace{-2mm}
\centering
\subfloat[\label{fig:dis_sad_du}]{%
            \centering
		\includegraphics[width=0.48\linewidth, clip]{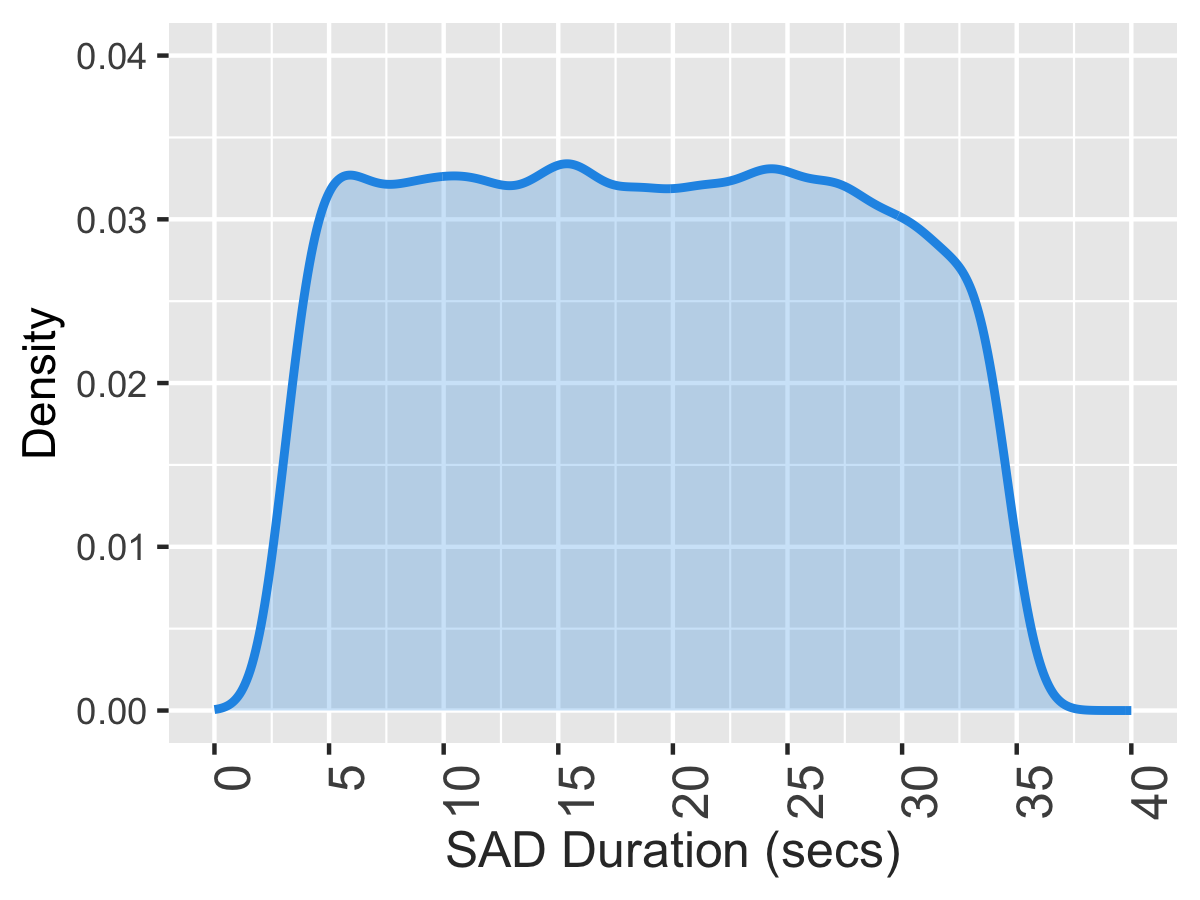}}
	\subfloat[\label{fig:t1_sad_du}]{%
            \centering
		\includegraphics[width=0.50\linewidth, clip]{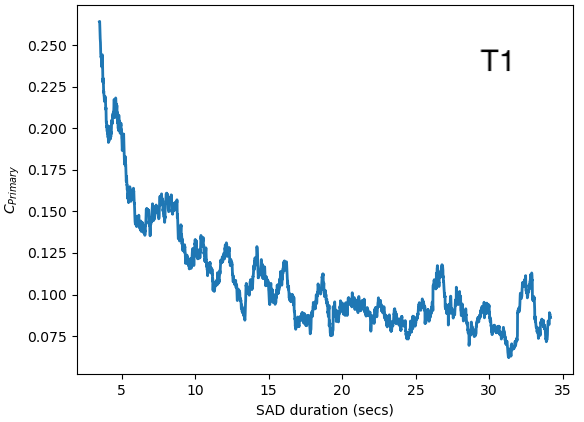}}\\
	\caption{\textit{SAD duration} effect on system performance (a) \textit{SAD duration} distribution for the \textit{dev} and \textit{test} set (b) T1 system performance vs. \textit{SAD duration}.}
 \vspace{-4mm}
\end{figure}

To examine language-pair confusability, we conducted data analysis using heatmap confusion matrices as shown in Figure~\ref{fig:confusion_heatmap}. The axes are language codes. The diagonal values from upper-left to bottom-right are $P_{Miss}$ (false reject rates) and the off-diagonal values are $P_{FA}$ (false alarm rates). A higher false alarm probability implies a potential confusability for that language pair. For simplicity, results of $P_{Target}=0.5$ for the four leading systems are demonstrated using heatmap confusion matrices. Given the \textit{test} set and systems, a higher confusability is observed for three clusters of language pairs as follows: 1) among Arabic languages (ara-aeb, ara-arq, ara-ayl), 2) between South African English (eng-ens) and Indian-accented South African English (eng-iaf), and 3) Ndebele (nbl-nbl), Tsonga (tso-tso), Venda (ven-ven), Xhosa (xho-xho) and Zulu (zul-zul). 




To gain insight on how metadata variables (i.e., factors) affect system performance, we conducted experiments given the metadata listed in Section~\ref{sec:data}.3. For simplicity, the following analyses are demonstrated using \textit{data source type} and  \textit{speech duration} only.
The LRE22 data was collected in two primary genres, namely, conversational telephone speech (CTS) and broadcast narrowband speech (BNBS) which we call \textit{data source type}. 
Figure~\ref{fig:partiton_data_source} shows system performance ($actC_{Primary}$) partitioned by \textit{data source type} (CTS vs BNBS) for all the \textit{primary} submissions under the \textit{fixed} training condition.
The top-left pie chart is a distribution of CTS and BNBS on the \textit{test} set, which is imbalanced. The bar plot shows a performance comparison between CTS (blue) and BNBS (orange) across all the teams. The results indicates that, given the imbalanced distribution, CTS is more challenging and that \textit{data source type} has a strong effect on system performance; a similar trend is observed across the systems. 


Durations of \textit{test} set segments varied between 3s and 35s of speech that have been randomly sampled and extracted from longer recordings as determined by an automatic Speech Activity Detector (SAD) which we call \textit{SAD duration}. 
Figure~\ref{fig:dis_sad_du} shows a distribution of \textit{SAD duration} for the \textit{test} set and Figures~\ref{fig:t1_sad_du} shows the performance of a top-performing system by \textit{SAD duration}. Given the \textit{test} set and systems, it is seen that when \textit{SAD duration} increases, $actC_{Primary}$ significantly decreases up to a certain duration (between 15s and 20s). After that, a diminishing return on system performance improvement is observed across the systems. 


\section{Conclusions}

We presented a summary of the 2022 NIST Language Recognition Evaluation with an emphasis on low resource languages and random duration of speech segments. 


The results showed that almost no calibration error was observed for the top-performing systems for both the \textit{fixed} and \textit{open} training condition. Overall, the submissions under the \textit{open} training condition had better performance compared to the \textit{fixed} condition submissions, with only one exception. Given the \textit{test} set and \textit{primary} systems under the \textit{fixed} training condition, we found that Oromo and Tigrinya were easier to detect while Xhosa and Zulu were harder to detect. A greater confusability was observed for the language pairs 1) among Zulu, Xhosa, Ndebele, Tsonga, and Venda, 2) between South African and Indian-accent South African English, and 3) among Tunisian, Algerian, and Libyan Arabic languages. Some of the metadata, such as \textit{data source type} and \textit{SAD duration}, had a significant effect on system performance for all systems. In terms of \textit{SAD duration}, when speech duration increased, system performance significantly increased up to a certain duration, and then we observed a diminishing return on system performance afterward. 

\section{Disclaimer}
\anonymize{These results presented in this paper are not to be construed or represented as endorsements of any participant's system, methods, or commercial product, or as official findings on the part of NIST or the U.S. Government.

The work of MIT Lincoln Laboratory (MITLL) is sponsored by the Department of Defense under Air Force Contract No. FA8702-15-D-0001. Any opinions, findings, conclusions or recommendations expressed in this material are those of the author(s) and do not necessarily reflect the views of the U.S. Air Force.} 



\break
\balance

\bibliographystyle{IEEEtran}

\bibliography{mybib}


\end{document}